\documentclass[runningheads]{llncs}
\usepackage[T1]{fontenc}
\usepackage{graphicx}
\usepackage{booktabs}
\usepackage[misc]{ifsym}
\newcommand{\corr}{(\Letter)}
\usepackage{subcaption}
\usepackage{xcolor}

\usepackage{soul}
\usepackage{url}
\usepackage[hidelinks]{hyperref}
\usepackage[utf8]{inputenc}
\usepackage{caption}
\usepackage{amsmath}
\usepackage{amssymb}
\usepackage{algorithm}
\usepackage{algorithmic}
\usepackage[switch]{lineno}
\usepackage{multirow}

\begin{document}

\title{PRISM: Powerful Time Series to Image (TS2I) Representations for Multivariate Anomaly Detection}

\toctitle{PRISM: Powerful Time Series to Image (TS2I) Representations for Multivariate Anomaly Detection}

\titlerunning{PRISM: Powerful TS2I Representations for Multivariate TSAD}

\author{Mateusz Smendowski\inst{1} \and
Kamil Faber\inst{1} \and
Piotr Nawrocki\inst{1} \and
Nathalie Japkowicz\inst{2} \and
Roberto Corizzo\inst{2}\corr}

\authorrunning{M. Smendowski et al.}

\tocauthor{Mateusz Smendowski, Kamil Faber, Piotr Nawrocki, Nathalie Japkowicz,  Roberto Corizzo}

\institute{Faculty of Computer Science, AGH University of Krakow,
al. A. Mickiewicza 30, 30-059 Krakow, Poland
\email{\{smendowski,kfaber,piotr.nawrocki\}@agh.edu.pl}
\and
Department of Computer Science, American University,
4400 Massachusetts Ave NW, Washington, DC 20016, USA
\email{\{japkowic,rcorizzo\}@american.edu}}
\maketitle              

\begin{abstract}
Time-series anomaly detection (TSAD) underpins applications in predictive maintenance, finance, and cloud computing, however performance remains sensitive to representation choices, especially in multivariate settings. While transforming time series into images has shown success in forecasting and classification, it remains unclear how multivariate, high-dimensional series should be mapped to multi-channel images and whether vision backbones can match time-domain baselines in TSAD. We introduce PRISM, a plug-and-play meta-workflow enabling systematic construction and evaluation of image-based representations for multivariate TSAD. Our evaluation spanning over 7,000 experiments shows that well-designed PRISM configurations are competitive with 24 time-domain baselines, achieving the best VUS-PR on 10 of 14 datasets, with an average improvement of 41\% over the best competing method on those datasets. Further, we identify channelization -- how the channel dimension of multi-channel images is constructed -- as a critical and previously understudied design dimension, and introduce MSM, a novel statistics-based scheme achieving 11–27\% gains over PCA-based alternatives. Finally, ImageNet-pretrained encoders transfer effectively to TSAD, with frozen encoders retaining 92\% of fine-tuned performance while training 1.8 times faster. Our code is available at: \url{https://github.com/Smendowski/PRISM}.

\keywords{
Anomaly detection, Multivariate time series, Time series to image representations, Transfer learning, Autoencoders, Computer vision
}
\end{abstract}

\section{Introduction}

Time-series anomaly detection (TSAD) is a core building block in applications such as predictive maintenance, finance, cybersecurity, cloud computing, and healthcare \cite{zamanzadeh2024deep}. Despite its broad applicability, TSAD remains a challenging problem due to the diversity of temporal dynamics and the strong dependence of detection performance on data representation choices, particularly in multivariate settings \cite{xi2025randomad,nawrocki2025survey}. Several factors motivate the transition from the time-series domain to the vision domain, including \romannumeral 1) encoding temporal and cross-variate structure spatially in a form compatible with highly optimized, pretrained computer vision (CV) backbones \cite{kang2025memory}; and \romannumeral 2) revealing structural patterns that are not directly accessible in one-dimensional signals \cite{ijcai2025p1178}.



In this paper, we address these gaps by introducing PRISM (\textbf{P}owe\textbf{R}ful t\textbf{I}me \textbf{S}eries to image representations for \textbf{M}ultivariate anomaly detection), a unified meta-workflow that enables multivariate TSAD through image-based representations. PRISM introduces the notion of a projection scheme as an end-to-end mapping from a multivariate time-series window to a fixed-size image. 

\begin{figure}[h!]
\centering
\includegraphics[width=\textwidth]{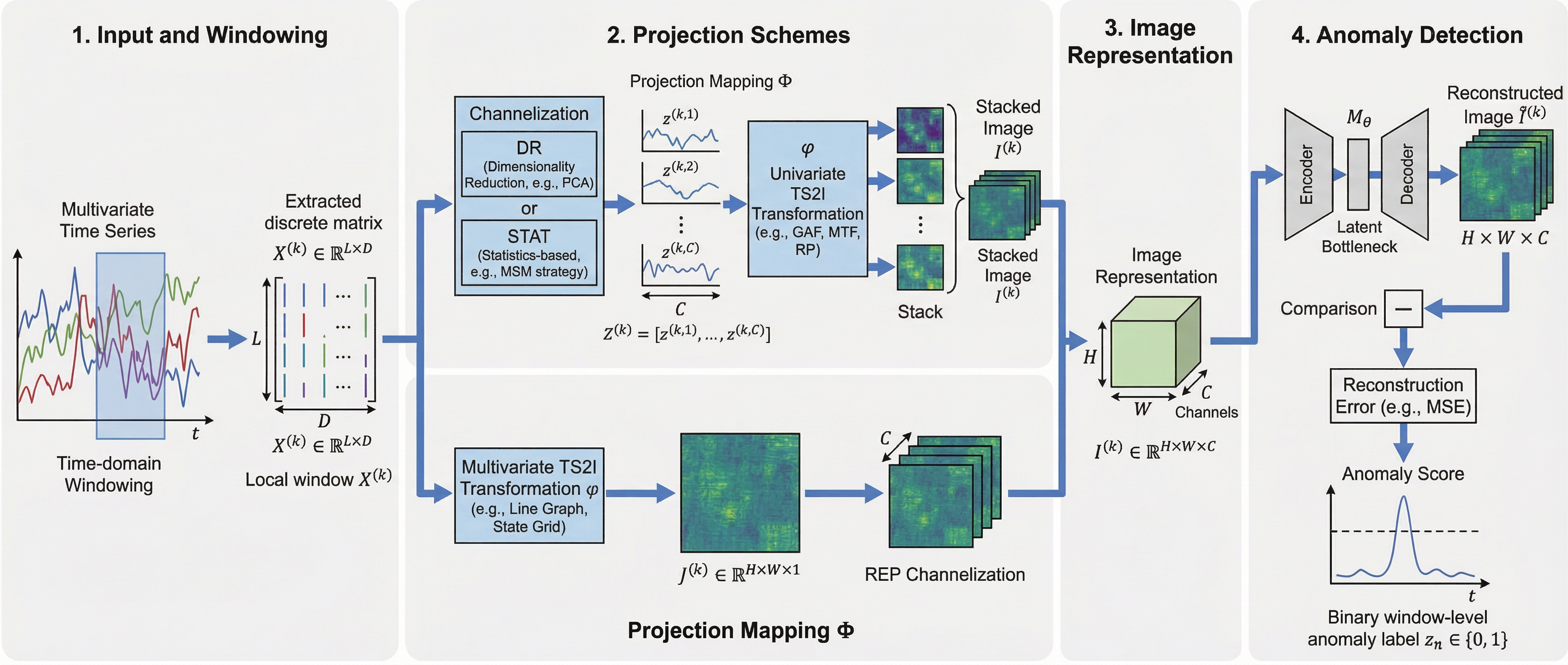}
\caption{Architecture of the proposed PRISM meta-workflow.}
\label{fig:main-figure}
\end{figure}

PRISM distinguishes two components of a projection scheme (Figure \ref{fig:main-figure}). The TS2I transformation specifies how a time-series window is mapped to a single-channel image, while channelization defines how the channel dimension is constructed to obtain a multi-channel representation for vision backbones. 
Crucially, PRISM is not a single anomaly detection method, but a plug-and-play meta-workflow that decouples representation design from detector choice. 
%




Our main contributions are summarized as follows:

\noindent - \textbf{C1}: We introduce PRISM, a unified plug-and-play meta-workflow -- a modular pipeline that decouples TS2I representation design from detector choice -- enabling systematic, reproducible construction and evaluation of image-based representations for multivariate TSAD with any autoencoder-based vision backbone or anomaly detector.

\noindent - \textbf{C2}: We identify channelization -- how the channel dimension of multi-channel images is constructed from multivariate time series -- as a critical and understudied design dimension in TS2I pipelines. We propose a taxonomy of channelization families and introduce MSM (Mean-Std-Max), a projection-free, statistics-based scheme specifically designed for TSAD, achieving 11–27\% VUS-PR gains over PCA-based alternatives across all tested configurations.

\noindent - \textbf{C3}: Through comprehensive evaluation spanning over 7,000 experiments across 15 TS2I projection schemes, 2 autoencoder architectures, and 14 datasets from the TSB-AD benchmark, we demonstrate that well-designed PRISM configurations are competitive with 24 specialized time-domain baselines, achieving the best VUS-PR on 10 of 14 datasets, with a mean relative VUS-PR improvement of 41\% across those datasets and a 6\% overall mean improvement, establishing TS2I representations as a competitive alternative to time-domain methods.

\noindent - \textbf{C4}: We demonstrate that ImageNet-pretrained encoders transfer effectively to TSAD despite the substantial domain gap, showing that frozen encoders retain 92\% of fully fine-tuned performance while training 1.8 times faster, and that MSM channelization is most critical when encoder adaptation is constrained.

\noindent - \textbf{C5}: We provide a computational cost analysis of PRISM, showing that TS2I transformation overhead represents only 2–8\% of total pipeline time, and identifying MWT-MSM and SG-REP as suitable configurations for accuracy-critical and latency-critical deployments respectively. 







\section{Related Work}

\noindent\textbf{Time series to image transformations (TS2I):}
TS2I include different approaches with limited applicability to multivariate data. Widely used univariate-only methods, such as Gramian Angular Summation Field (GASF), Gramian Angular Difference Field (GADF), Markov Transition Field (MTF), and Recurrence Plot (RP)~\cite{10.5555/2832747.2832798}, encode temporal patterns through polar coordinates, state transitions, or recurrence structures, becoming standard baselines in forecasting and classification, yet GASF, GADF, and RP exhibit $\mathcal{O}(n^2)$ complexity and produce symmetric heatmaps unsuitable for high-dimensional settings ~\cite{ZHANG2021102537}, while MTF, though sharing the quadratic scaling, encodes state transition probabilities and does not produce symmetric outputs.
Line plots provide direct visual encodings with demonstrated classification performance on univariate time series across 98 UCR datasets~\cite{rodrigues2021plotting,yang2023your}. For multivariate data, grid line plots extend this approach by arranging multiple trajectories within a single image but incur overhead due to grid rendering~\cite{NEURIPS2023_9a17c1eb}.
Alternative approaches operate in the frequency domain, including STFT spectrograms and specialized wavelets (Ricker, Morlet), while similarity-based methods employ DTW matrices and shapelet maps~\cite{ijcai2025p1178}. Few TS2I transformations explicitly target multivariate time series encoding. State Grid discretizes time series data into bins \cite{karami2025timehr}. Separately, correlation-based feature reordering achieves $\mathcal{O}(n)$ complexity but has been applied primarily to forecasting rather than anomaly detection \cite{kang2025memory}.
Despite this diversity, key problems remain: it is unclear how multivariate, high-dimensional time series should be effectively turned into images~\cite{ijcai2025p1178}, and more importantly, how multiple channels should be organized in the image. PRISM addresses this by establishing complete encoding workflows for multivariate TSAD and providing a comprehensive performance landscape for image-based TSAD: systematic evaluation of 6 univariate and 3 multivariate TS2I transformations combined with 3 channelization strategies across 14 benchmark datasets.

\noindent\textbf{Time Series Anomaly Detection:}
TSAD has traditionally relied on statistical methods, but machine learning approaches are becoming more common~\cite{boniol2024dive}. Current methods span distance-based, density-based, prediction-based, and reconstruction-based approaches~\cite{namura2024training}, with complex deep learning architectures such as autoencoder-based detectors~\cite{sakurada2014anomaly,faber2025xlstmad}. Recent advances include transformer-based reconstruction with trend mechanisms~\cite{ijcai-liu2025rtdetector}, LLM knowledge distillation~\cite{ijcai2024p239}, and contrastive learning~\cite{tang2024perturbation}. Despite these advances, time-domain methods remain sensitive to representation choices and often struggle to scale to high-dimensional multivariate settings, motivating the exploration of alternative representation paradigms.
A separate group of methods transforms time series into images, enabling the application of vision models to anomaly detection. One strategy applies TS2I transformations across the variable dimension rather than time~\cite{MA2024102232}, encoding inter-variable correlations but limiting temporal modeling. Alternative approaches generate separate per-variable images using GAF, RP, and MTF with adversarial autoencoders~\cite{kang2024time} or temporal stacking with CNNs~\cite{sid-ryan-icmla-1:2019}, producing single-channel outputs that are incompatible with pretrained RGB vision backbones, and are evaluated only on limited benchmarks.
In contrast, PRISM models both temporal dynamics and cross-variate structure by transforming multivariate windows into multi-channel images via principled channelization. We provide thorough evaluation across 14 TSB-AD datasets, 9 challenging metrics, and 24 time-domain baselines, addressing scalability to high-dimensional data, resource efficiency, and pretrained backbone compatibility, providing a comprehensive empirical landscape for image-based multivariate TSAD and establishing PRISM as a plug-and-play framework compatible with any choice of vision backbone or detector.

\noindent\textbf{Transfer learning:} 
Transfer learning (TL) has become a cornerstone of modern computer vision (CV), enabling practitioners to leverage pretrained backbones -- ranging from CNNs (ResNet, AlexNet, EfficientNet, MobileNet) to transformers (ViT, PVT, Swin)~\cite{10508091} -- rather than training from scratch. This approach reduces data and computational requirements~\cite{yan2024comprehensive}. However, TL success can be sensitive to domain similarity; a key open question is whether TS2I representations are sufficiently image-like to benefit from ImageNet pretraining. Recent work has also explored training-free anomaly detection using pretrained vision models like PatchCore on hybrid TS2I representations combining wavelets (Ricker, Morlet) with frequency embeddings~\cite{namura2024training}, eliminating task-specific training entirely. While promising, this approach forgoes the potential benefits of fine-tuning to the target domain.
Existing studies~\cite{ecml} evaluate only a few TL strategies on limited datasets, and the feasibility of positive cross-domain transfer remains poorly understood. We address this by evaluating whether an ImageNet-pretrained AE with ResNet-based encoder can achieve positive transfer to TS2I representations, systematically comparing three adaptation strategies across 14 datasets from the TSB-AD benchmark.

\section{PRISM}

\subsection{Preliminaries}

Let $X \in \mathbb{R}^{T \times D}$ denote a multivariate time series with $T$ time steps
and $D$ dimensions. The observation at time $t$ is a vector $x_t \in \mathbb{R}^{D}$, so that $X = [x_1;\dots;x_T].$
In the supervised setting, we denote by $y_t \in \{0,1\}$ the point-wise anomaly label at time $t$, where $y_t = 1$ indicates an anomalous observation. 

\noindent\textbf{Time-domain windowing:}
Given a window length $L$ and stride $s$, the multivariate time series is segmented in the time domain into a sequence of fixed-length windows: 
\[
    X^{(k)} = (x_{t_k}, \dots, x_{t_k+L - 1})
        \in \mathbb{R}^{L \times D},
    \qquad
    t_k = 1+(k-1)s.
\]


%
\noindent For evaluation purposes, a window-level anomaly label $z_k \in \{0,1\}$ is derived via: $z_k = \max_{t \in \{t_k, \dots, t_k+L - 1\}} y_t.$


\noindent\textbf{Workflows for projection schemes:}
In our work, we define a projection scheme as a process that converts time series to the image domain. 
Each window $X^{(k)}$ is mapped to a fixed-size image $I^{(k)}$ 
through a projection scheme $\Phi$, and formally defined as
\[
    I^{(k)} = \Phi\!\big(X^{(k)}\big),
    \qquad
    \Phi : \mathbb{R}^{L \times D} \rightarrow \mathbb{R}^{H \times W \times C}.
\]

\noindent Here, $H$ and $W$ denote the spatial resolution of the resulting image, and $C$ denotes the number of channels. PRISM distinguishes two families of projection schemes, both of which are applied to multivariate data and differ in the way the window $X^{(k)}$ is projected to the image domain. 




\subsection{Projection scheme for univariate-based TS2I transformations}

\label{sec:univar-ts2i-projection-scheme}
Since many TS2I transformations can only work with univariate data, we devise an approach to adapt them for multivariate data, which is the main focus of our study.
%
Specifically, methods such as GAF, MTF, RP require a one-dimensional time series and therefore cannot be applied directly to multivariate windows $X^{(k)} \in \mathbb{R}^{L \times D}$. 

\noindent In this scheme, the projection mapping $\Phi$ consists of a channelization step that derives a set of univariate series from the multivariate window, followed by the application of a univariate TS2I transformation $\varphi$ to each resulting channel.

\subsubsection{DR channelization}
In dimensionality reduction-based (DR) channelization, the multivariate window $X^{(k)} \in \mathbb{R}^{L \times D}$ is projected onto a $C$-dimensional (where $C$ is lower than original dimensionality) subspace along the variable dimension using DR techniques such as PCA.
This yields $C$ univariate series defined as:
\[
    Z^{(k)} = \big[z^{(k,1)}, \dots, z^{(k,C)}\big],
    \qquad
    z^{(k,c)} \in \mathbb{R}^{L}.
\]

\noindent Each derived series is independently transformed into a single-channel image using the same univariate TS2I transformation $\varphi$:
\[
    J^{(k,c)} = \varphi\!\big(z^{(k,c)}\big),
    \qquad
    J^{(k,c)} \in \mathbb{R}^{H \times W},
    \quad c = 1,\dots,C.
\]

\noindent The final image representation is obtained by stacking the resulting images along the channel dimension:
\[
    I^{(k)} =
    \text{stack}\big(J^{(k,1)}, \dots, J^{(k,C)}\big)
    \in \mathbb{R}^{H \times W \times C}.
\]

\subsubsection{STAT channelization}
In addition to DR channelization, we devise an original statistics-based (STAT) channelization approach for TSAD, where channels are constructed directly from the multivariate window $X^{(k)} \in \mathbb{R}^{L \times D}$ using statistical descriptors computed across the variable dimension. One advantage is that this approach is projection-free, avoiding intermediate latent representations. 

\noindent Formally, a set of $C$ statistical operators is applied to obtain derived univariate series
\[
    Z^{(k)} = \big[z^{(k,1)}, \dots, z^{(k,C)}\big],
    \qquad
    z^{(k,c)} \in \mathbb{R}^{L},
\]
where each $z^{(k,c)}$ corresponds to a separate statistic evaluated at each time step.
%
Each derived series is independently transformed into a single-channel image using the same univariate TS2I transformation $\varphi$:
\[
    J^{(k,c)} = \varphi\!\big(z^{(k,c)}\big),
    \qquad
    J^{(k,c)} \in \mathbb{R}^{H \times W},
    \quad c = 1,\dots,C.
\]

\noindent The final image representation is constructed by stacking the resulting images along the channel dimension:
\[
    I^{(k)} =
    \text{stack}\big(J^{(k,1)}, \dots, J^{(k,C)}\big)
    \in \mathbb{R}^{H \times W \times C}.
\]

\paragraph{MSM strategy}
Although STAT channelization is general and supports any statistics and time series dimensionality, we devise a novel MSM (Mean-Std-Max) strategy that highlights salient statistical properties for TSAD. The three channels are chosen for their complementary sensitivity to distinct anomaly types: the mean captures global level shifts across variables, the standard deviation encodes changes in 
cross-variate spread or synchrony, and the maximum preserves extreme individual deviations that averaging would otherwise dilute. Together, they characterize the first-order cross-variate distributional shape at each time step without assuming any covariance structure -- making MSM sensitive to a broader class of anomalies than PCA-based channelization, which maximizes variance explained in normal data 
rather than discriminability between normal and anomalous patterns. MSM's projection-free nature further confers robustness to the non-stationarity common in real-world multivariate time series, as empirically validated in Section \ref{sec:channelization}.


\subsection{Projection scheme for multivariate-based TS2I transformations}
\label{sec:multivar-ts2i-projection-scheme}

The projection scheme for multivariate-based TS2I transformations applies to transformation methods that operate natively on multivariate inputs. Such methods (e.g., Line Graph or State Grid) directly process a window $X^{(k)} \in \mathbb{R}^{L \times D}$ and support arbitrary input dimensionality.
A concrete multivariate TS2I transformation, denoted by $\varphi$, is applied to obtain a single-channel image:
\[
    J^{(k)} = \varphi\!\big(X^{(k)}\big),
    \qquad
    J^{(k)} \in \mathbb{R}^{H \times W \times 1}.
\]
The overall projection scheme $\Phi$ comprises this transformation followed by a channelization strategy.

\subsubsection{REP channelization}

Given $J^{(k)} \in \mathbb{R}^{H \times W}$, replication-based (REP) channelization constructs the final image by replicating the single-channel output across the channel dimension. While this introduces no additional information across channels, it serves two practical purposes: satisfying the 3-channel input requirement of pretrained RGB vision backbones, and enabling a controlled baseline that isolates the contribution of the TS2I transformation itself from any channelization effect. Formally, for a target number of channels $C$, the representation is given by
\[
    I^{(k)} =
    \text{stack}\big(\underbrace{J^{(k)}, \dots, J^{(k)}}_{C \text{ times}}\big)
    \in \mathbb{R}^{H \times W \times C}.
\]

\subsection{Anomaly Detection}

In PRISM, 
an autoencoder-based anomaly detection model $M_{\theta} : \mathbb{R}^{H \times W \times C} \rightarrow \mathbb{R}$ is trained to assign an anomaly score (i.e., reconstruction error) to each image-based window representation, where $\theta$ denotes model parameters. 
PRISM's plug-and-play design allows any anomaly detection model to serve as the scoring component. In this study, we adopt autoencoder-based architectures due to their powerful non-linear modeling capabilities, widely recognized as a standard approach in the current literature. 



\section{Experimental Setup}
In our experiments, we leverage the TSB-AD benchmark and follow its standardized protocol \cite{NEURIPS2024_c3f3c690}. For reproducibility, we share our code at \url{https://github.com/Smendowski/PRISM}.

Framework-level ablation is constrained by strict sequential dependencies (e.g., the projection scheme cannot be omitted because subsequent pipeline stages fundamentally rely on its image output representation). To assess component contributions, we exhaustively evaluate configurations of various TS2I feature extraction methods, channelization strategies, and predictive models. Our extensive evaluation resulted in more than 7,000 training and evaluation runs, all performed on NVIDIA GH200 GPUs. 

\subsection{Datasets and metrics}

In this study, we focus on the following 14 multivariate public datasets: CATSv2 (dimensionality $D=17$), CreditCard ($D=29$), Daphnet ($D=9$), Exathlon ($D=21$), GECCO ($D=9$), Genesis ($D=18$), LTDB ($D=2$), MSL ($D=55$), OPP ($D=248$), PSM ($D=25$), SMAP ($D=25$), SMD ($D=38$), SWaT ($D=59$), and TAO ($D=3$). They are available at \url{https://github.com/TheDatumOrg/TSB-AD} and their detailed descriptions can be found in the work introducing TSB-AD benchmark~\cite{NEURIPS2024_c3f3c690}. 


We follow the TSB-AD evaluation protocol, reporting metrics across three categories: ranking-based (AUC-PR, AUC-ROC, VUS-PR, VUS-ROC), point-wise (Standard -- S-F1, Point-Adjust -- PA-F1), and event-level (Event-based -- E-F1, Range-based -- R-F1, Affiliation -- A-F1), jointly assessing detector quality across thresholds, individual timestamps, and contiguous anomaly segments.

\subsection{TS2I projection schemes}

In PRISM, we employ a diverse set of state-of-the-art TS2I transformations to map multivariate time-series windows into fixed-size images. We set the window length to $L = 20$ and stride $s = 1$ across all datasets to ensure a controlled, fair comparison across the 14 benchmark datasets -- varying window length per dataset would introduce a confounding factor that conflates representation quality with hyperparameter tuning. Moreover, in TSB-AD, many methods use a single window length across all datasets. We acknowledge that for datasets with longer anomaly patterns or slower dynamics (e.g., OPP with $D = 248$), a fixed short window may limit the capture of long-range dependencies. A data-driven window selection approach -- for instance, based on autocorrelation analysis to identify the dominant periodicity of each series -- could unlock further gains.

The univariate-based TS2I transformations considered in this study include Gramian Angular Summation Field (GASF), Gramian Angular Difference Field (GADF), Markov Transition Field (MTF), Recurrence Plot (RP), Morlet Wavelet Transform (MWT), and Ricker Wavelet Transform (RWT). These methods are employed within the projection scheme for univariate-based TS2I transformations, where two channelization variants are evaluated. 

As a representative of DR channelization, we adapt PCA with $C = 3$ principal components. 
In addition, we evaluate our original three-channel MSM scheme. 
%
We also consider multivariate-based TS2I transformations, such as Random Noise (RN), State Grid (SG), and Line Graph (LG), which use REP channelization within the projection scheme for multivariate-based TS2I transformations.

\begin{figure}[h!]
\centering

\begin{tabular}{ccccc}
\subcaptionbox{RN-REP}{\includegraphics[width=0.155\textwidth]{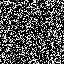}} &
\subcaptionbox{SG-REP}{\includegraphics[width=0.155\textwidth]{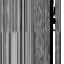}} &
\subcaptionbox{LG-REP}{\includegraphics[width=0.155\textwidth]{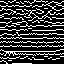}} &
\subcaptionbox{GASF-PCA}{\includegraphics[width=0.155\textwidth]{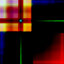}} &
\subcaptionbox{GASF-MSM}{\includegraphics[width=0.155\textwidth]{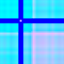}}
\\

\subcaptionbox{GADF-PCA}{\includegraphics[width=0.155\textwidth]{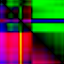}} &
\subcaptionbox{GADF-MSM}{\includegraphics[width=0.155\textwidth]{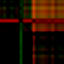}} &
\subcaptionbox{MTF-PCA}{\includegraphics[width=0.155\textwidth]{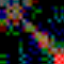}} &
\subcaptionbox{MTF-MSM}{\includegraphics[width=0.155\textwidth]{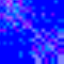}} &
\subcaptionbox{RP-PCA}{\includegraphics[width=0.155\textwidth]{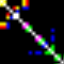}}
\\

\subcaptionbox{RP-MSM}{\includegraphics[width=0.155\textwidth]{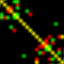}} &
\subcaptionbox{MWT-PCA}{\includegraphics[width=0.155\textwidth]{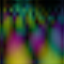}} &
\subcaptionbox{MWT-MSM}{\includegraphics[width=0.155\textwidth]{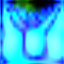}} &
\subcaptionbox{RWT-PCA}{\includegraphics[width=0.155\textwidth]{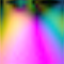}} &
\subcaptionbox{RWT-MSM}{\includegraphics[width=0.155\textwidth]{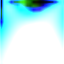}} \\




\end{tabular}
\caption{
Multi-channel image representations of a single multivariate time series window across different TS2I projection methods and channelization strategies. Each 64×64 RGB image encodes the same window differently, illustrating how the choice of projection and channelization affects the resulting visual structure.
}
\label{fig:tsi_overview}
\end{figure}

We set the image resolution to $H = W = 64$ and use $C = 3$ channels corresponding to the standard RGB format. This configuration provides a favorable trade-off between representational capacity and computational efficiency while enabling the application of pretrained backbones \cite{namura2024training}. In total, our experimental setup involves 15 TS2I projection schemes as illustrated in Figure~\ref{fig:tsi_overview}.

\subsection{Models}

In our study, we employ two vision backbones: \romannumeral 1) CNN-AE: a lightweight 2D convolutional autoencoder with three encoding layers, and \romannumeral 2) ResNet-AE: an autoencoder with a ResNet encoder initialized with ImageNet-pretrained weights and paired with a symmetric decoder. For ResNet-AE, we evaluate three TL strategies: Frozen Encoder (FE), which trains only the decoder while keeping the encoder frozen; Progressive Encoder Unfreeze (PEU), which starts with a frozen encoder and fully unfreezes it mid-training with a reduced learning rate; and Differential Learning Rates (DLR), which trains encoder and decoder simultaneously from the start using different learning rates. All architectural details and hyperparameters are documented in our repository at \url{https://github.com/Smendowski/PRISM}.

To evaluate PRISM, we compare against 24 time-domain baselines including recent xLSTMAD~\cite{faber2025xlstmad} and 23 additional methods from the TSB-AD benchmark: CNN, OmniAnomaly, PCA, LSTMAD, USAD, AutoEncoder, KMeansAD, CBLOF, MCD, OCSVM, Donut, RobustPCA, FITS, OFA, EIF, COPOD, IForest, HBOS, TimesNet, KNN, TranAD, LOF, and AnomalyTransformer.

\section{Experimental results}
\begin{table*}[h!]
\scriptsize
\centering
\caption{VUS-PR of representative detection approaches on 14 TSB-AD-M datasets. PRISM is built upon CNN and RN18 (ResNet18) auto-encoder backbones. Best results are shown in \textbf{bold}, while second-best results are \underline{underlined}. Only the 10 best-performing PRISM configurations and 10 best-performing non-PRISM approaches are reported in the table (20 out of 54 in total); additional results are available at {\color{blue}\url{https://github.com/Smendowski/PRISM}}.}
\scriptsize
\label{tab:vuspr}
\begin{tabular}{|l|l|c|c|c|c|c|c|c|c|c|c|c|c|c|c|c}
\hline
& \textbf{Method/Config.} &
\makebox[0pt][c]{\rotatebox{90}{\textbf{CATSv2}}} &
\makebox[0pt][c]{\rotatebox{90}{\textbf{CreditCard }}} &
\makebox[0pt][c]{\rotatebox{90}{\textbf{Daphnet}}} &
\makebox[0pt][c]{\rotatebox{90}{\textbf{Exathlon}}} &
\makebox[0pt][c]{\rotatebox{90}{\textbf{GECCO}}} &
\makebox[0pt][c]{\rotatebox{90}{\textbf{Genesis}}} &
\makebox[0pt][c]{\rotatebox{90}{\textbf{LTDB}}} &
\makebox[0pt][c]{\rotatebox{90}{\textbf{MSL}}} &
\makebox[0pt][c]{\rotatebox{90}{\textbf{OPP}}} &
\makebox[0pt][c]{\rotatebox{90}{\textbf{PSM}}} &
\makebox[0pt][c]{\rotatebox{90}{\textbf{SMAP}}} &
\makebox[0pt][c]{\rotatebox{90}{\textbf{SMD}}} &
\makebox[0pt][c]{\rotatebox{90}{\textbf{SWaT}}} &
\makebox[0pt][c]{\rotatebox{90}{\textbf{TAO}}} \\
\hline

\multirow{10}{*}{\rotatebox{90}{\quad \textbf{PRISM} \quad}}
& CNN+SG-REP     & 0.11 & \textbf{0.20} & \underline{0.43} & 0.86 & \underline{0.37} & 0.06 & 0.28 & 0.39 & 0.17 & 0.17 & 0.22 & 0.45 & 0.33 & 0.95 \\
& CNN+MWT-MSM    & \underline{0.23} & 0.16 & 0.35 & 0.86 & 0.13 & 0.45 & 0.29 & \underline{0.45} & 0.13 & 0.28 & 0.28 & \underline{0.46} & \underline{0.60} & 0.91 \\

& RN18+SG-REP & 0.10 & 0.17 & 0.35 & \underline{0.98} & 0.31 & \underline{0.80} & 0.38 & 0.43 & 0.21 & 0.15 & 0.32 & 0.44 & \textbf{0.64} & 0.89 \\
& RN18+LG-REP     & 0.06 & 0.14 & 0.45 & 0.98 & \textbf{0.43} & 0.74 & 0.28 & 0.23 & \textbf{0.39} & 0.25 & 0.24 & 0.40 & 0.48 & 1.00 \\
& RN18+GASF-PCA   & 0.12 & 0.03 & 0.29 & 0.29 & 0.20 & 0.14 & 0.37 & 0.23 & 0.06 & 0.18 & 0.22 & 0.18 & 0.32 & \underline{1.00} \\
& RN18+GASF-MSM   & 0.10 & \underline{0.19} & 0.15 & 0.74 & 0.11 & 0.19 & 0.30 & 0.24 & 0.06 & 0.16 & 0.24 & 0.15 & 0.18 & \textbf{1.00} \\
& RN18+MWT-PCA    & 0.10 & 0.06 & \textbf{0.47} & 0.97 & 0.09 & 0.16 & 0.21 & 0.27 & 0.20 & 0.28 & 0.25 & 0.45 & 0.49 & 0.99 \\
& RN18+MWT-MSM    & \textbf{0.25} & 0.13 & 0.41 & 0.86 & 0.12 & 0.51 & 0.37 & \textbf{0.48} & 0.13 & 0.31 & \underline{0.35} & \textbf{0.48} & 0.59 & 0.93 \\
& RN18+RWT-PCA    & 0.09 & 0.05 & 0.38 & \textbf{0.98} & 0.10 & 0.46 & 0.20 & 0.29 & 0.17 & 0.27 & 0.29 & 0.39 & 0.28 & 0.81 \\
& RN18+RWT-MSM    & 0.18 & 0.16 & 0.35 & 0.86 & 0.16 & 0.60 & \underline{0.42} & 0.42 & 0.14 & \underline{0.34} & 0.34 & 0.45 & 0.59 & 0.86 \\
\hline
\multirow{10}{*}{\rotatebox{90}{\quad \textbf{Baselines} \quad}} 
& xLSTMAD            & \underline{0.23} & 0.06 & 0.28 & 0.94 & 0.19 & 0.67 & 0.42 & 0.33  & \underline{0.31} & \textbf{0.40} & 0.27 & 0.40 & 0.52 & 0.73 \\
& CNN                & 0.08 & 0.02 & 0.21 & 0.68 & 0.03 & 0.10 & 0.33 & 0.35 & 0.16 & 0.22 & 0.19 & 0.35 & 0.41 & 1.00 \\
& OmniAnomaly        & 0.04 & 0.02 & 0.34 & 0.84 & 0.02 & 0.00 & \textbf{0.44} & 0.22 & 0.18 & 0.16 & 0.12 & 0.17 & 0.15 & 0.81 \\
& PCA                & 0.12 & 0.10 & 0.13 & 0.95 & 0.20 & 0.02 & 0.24 & 0.15 & 0.30 & 0.16 & 0.09 & 0.36 & 0.45 & 1.00 \\
& LSTMAD             & 0.04 & 0.02 & 0.31 & 0.82 & 0.02 & 0.04 & 0.30 & 0.22 & 0.17 & 0.24 & 0.16 & 0.33 & 0.16 & 0.99 \\
& USAD               & 0.04 & 0.02 & 0.34 & 0.84 & 0.02 & 0.00 & 0.41 & 0.23 & 0.18 & 0.19 & 0.11 & 0.16 & 0.15 & 0.81 \\
& AutoEncoder        & 0.06 & 0.03 & 0.13 & 0.91 & 0.05 & 0.01 & 0.21 & 0.22 & 0.14 & 0.28 & 0.13 & 0.30 & 0.58 & 1.00 \\
& KMeansAD           & 0.12 & 0.02 & 0.30 & 0.37 & 0.06 & \textbf{0.89} & 0.41 & 0.44 & 0.06 & 0.21 & \textbf{0.38} & 0.36 & 0.16 & 0.86 \\
& MCD                & 0.13 & 0.06 & 0.14 & 0.80 & 0.03 & 0.06 & 0.21 & 0.23 & 0.17 & 0.26 & 0.10 & 0.26 & 0.54 & 1.00 \\
& Donut              & 0.07 & 0.02 & 0.17 & 0.66 & 0.03 & 0.18 & 0.26 & 0.30 & 0.15 & 0.20 & 0.18 & 0.19 & 0.44 & 0.75 \\
\hline
\end{tabular}
\end{table*}
This research is directed at answering the following research questions:

\noindent - \textbf{RQ1}: Can vision-based approaches using TS2I representations achieve competitive performance against state-of-the-art time-domain TSAD baselines?

\noindent - \textbf{RQ2}: How do different TS2I projection schemes balance detection performance and computational efficiency in multivariate TSAD?

\noindent - \textbf{RQ3}: What is the relative importance of channelization strategy versus autoencoder architecture in determining TSAD performance?

\noindent - \textbf{RQ4}: Can transfer learning from general vision domains (ImageNet) provide positive transfer to TSAD, and which encoder adaptation strategies are most effective?

\noindent - \textbf{RQ5}: What are the computational trade-offs between TS2I projection schemes and autoencoder architectures, and can the overhead introduced by data transformation be justified by superior TSAD performance?

\subsection{PRISM vs. time-domain baselines}
Table~\ref{tab:vuspr} compares PRISM variants against 24 established TSAD methods across 14 diverse multivariate datasets (dimensionality $D=2$ to $D=248$). 
Within the PRISM meta-workflow, top configurations achieved the best performance on 10 of 14 datasets, 
improving the results over the best non-PRISM method on average by 41\%, computed as the mean relative improvement of the best PRISM configuration over the best non-PRISM method across the 10 datasets where PRISM achieves highest performance.


Specifically, PRISM configurations achieved the highest VUS-PR improvements on the following datasets: GECCO by 115.0\% vs PCA (0.43 vs 0.20); CreditCard by 100.0\% vs PCA (0.20 vs 0.10); CATSv2 by 8.7\% vs xLSTMAD (0.25 vs 0.23); SMD by 20\% vs xLSTMAD (0.48 vs 0.40). On the other hand, time-series-based competitors achieved high performance on 2 lower-dimensional datasets: KMeansAD on Genesis ($D = 18$, 0.89 vs 0.80 for best PRISM configuration) and OmniAnomaly on LTDB ($D = 2$, 0.44 vs 0.42).


Across metrics, Table~\ref{tab:tsb-ad-m} presents mean performance across all 14 datasets and 9 evaluation metrics. 
PRISM 
achieved the best overall performance in 6 out of 9 metrics. Specifically, comparing it with the best non-PRISM method, it achieved the best VUS-PR (0.53 vs 0.50), AUC-PR (0.53 for PRISM vs 0.42 for non-PRISM), Standard F1 (0.58 vs 0.48), PA-F1 (0.96 vs 0.88), Event-based-F1 (0.86 vs 0.68), and R-based-F1 (0.45 vs 0.44).
%
%
%
Among baselines, only Donut was able to outperform PRISM and only in two metrics: VUS-ROC (0.86 vs 0.81 for PRISM) and Affiliation-F1 (0.98 vs 0.95 for PRISM). However, Donut lagged substantially on precision-recall metrics (AUC-PR: 0.24 vs 0.53).

%
The strong cross-dataset and cross-metric performance demonstrates that TS2I transformation with visual autoencoders offers a competitive dataset-agnostic framework, particularly effective for high-dimensional multivariate series where visual representations capture complex cross-channel dependencies that sequential architectures may struggle to model (\textbf{RQ1}). To streamline discussion, in the remainder of Section~5 we focus on VUS-PR; however, fine-grained results across all metrics are available in Table~\ref{tab:tsb-ad-m} and our repository.

\begin{table*}[h!]
\centering
\scriptsize
\caption{Mean performance across 9 evaluation metrics for 30 PRISM plug-and-play configurations and 24 time-domain baseline methods on 14 TSB-AD datasets. PRISM is built upon CNN and RN18 (ResNet18) auto-encoder backbones. Best results in \textbf{bold}, second-best \underline{underlined}.}
\label{tab:tsb-ad-m}
\begin{tabular}{|l|l|c|c|c|c|c|c|c|c|c|}
\hline
& \textbf{Method} & \textbf{AUC-} & \textbf{AUC-} & \textbf{VUS-} & \textbf{VUS-} & \textbf{S-F1} & \textbf{PA-F1} & \textbf{E-F1} & \textbf{R-F1} & \textbf{A-F1} \\
& & \textbf{PR} & \textbf{ROC} & \textbf{PR} & \textbf{ROC} & & & & & \\
\hline
\multirow{30}{*}{\rotatebox{90}{\quad \textbf{PRISM} \quad}}
& CNN+RN-REP     & 0.14 & 0.50 & 0.15 & 0.52 & 0.21 & 0.94 & 0.39 & 0.33 & 0.80 \\
&  CNN+SG-REP     & 0.49 & 0.75 & 0.48 & 0.76 & 0.53 & 0.90 & 0.79 & 0.42 & 0.93 \\
& CNN+LG-REP     & 0.45 & 0.73 & 0.44 & 0.75 & 0.50 & 0.93 & 0.76 & 0.40 & 0.92 \\
& CNN+GASF-PCA   & 0.24 & 0.61 & 0.23 & 0.63 & 0.31 & 0.94 & 0.62 & 0.37 & 0.88 \\
& CNN+GASF-MSM   & 0.38 & 0.70 & 0.37 & 0.71 & 0.43 & 0.93 & 0.73 & 0.36 & 0.90 \\
& CNN+GADF-PCA   & 0.27 & 0.53 & 0.26 & 0.55 & 0.34 & 0.88 & 0.64 & 0.40 & 0.89 \\
& CNN+GADF-MSM   & 0.26 & 0.54 & 0.25 & 0.55 & 0.37 & 0.90 & 0.62 & 0.42 & 0.86 \\
& CNN+MTF-PCA    & 0.24 & 0.57 & 0.24 & 0.59 & 0.30 & 0.93 & 0.63 & 0.39 & 0.86 \\
& CNN+MTF-MSM    & 0.35 & 0.63 & 0.35 & 0.64 & 0.42 & 0.92 & 0.66 & 0.39 & 0.88 \\
& CNN+RP-PCA     & 0.22 & 0.57 & 0.22 & 0.58 & 0.28 & 0.86 & 0.53 & 0.30 & 0.84 \\
& CNN+RP-MSM     & 0.31 & 0.62 & 0.31 & 0.63 & 0.37 & 0.83 & 0.59 & 0.30 & 0.88 \\
& CNN+MWT-PCA    & 0.49 & 0.76 & 0.48 & 0.77 & 0.53 & 0.89 & 0.77 & 0.41 & 0.93 \\
& CNN+MWT-MSM    & 0.51 & 0.74 & 0.51 & 0.75 & 0.56 & 0.92 & \underline{0.85} & \underline{0.45} & 0.94 \\
& CNN+RWT-PCA    & 0.45 & 0.72 & 0.45 & 0.73 & 0.51 & 0.90 & 0.75 & 0.40 & 0.92 \\
& CNN+RWT-MSM    & 0.47 & 0.74 & 0.47 & 0.75 & 0.54 & 0.92 & 0.82 & 0.42 & 0.93 \\

& RN18+RN-REP     & 0.14 & 0.50 & 0.15 & 0.52 & 0.21 & 0.93 & 0.32 & 0.32 & 0.77 \\
& RN18+SG-REP     & \underline{0.52} & 0.76 & \underline{0.52} & 0.77 & \underline{0.57} & 0.93 & 0.81 & 0.41 & 0.93 \\
& RN18+LG-REP     & 0.50 & \textbf{0.80} & 0.49 & 0.81 & 0.54 & \textbf{0.96} & 0.80 & 0.39 & 0.93 \\
& RN18+GASF-PCA   & 0.27 & 0.65 & 0.27 & 0.66 & 0.35 & \underline{0.96} & 0.70 & 0.37 & 0.90 \\
& RN18+GASF-MSM   & 0.38 & 0.73 & 0.37 & 0.74 & 0.44 & 0.94 & 0.77 & 0.36 & 0.92 \\
& RN18+GADF-PCA   & 0.30 & 0.62 & 0.29 & 0.63 & 0.36 & 0.94 & 0.67 & 0.37 & 0.89 \\
& RN18+GADF-MSM   & 0.30 & 0.58 & 0.29 & 0.60 & 0.39 & 0.94 & 0.67 & 0.38 & 0.88 \\
& RN18+MTF-PCA    & 0.23 & 0.58 & 0.23 & 0.59 & 0.30 & 0.93 & 0.60 & 0.37 & 0.86 \\
& RN18+MTF-MSM    & 0.33 & 0.63 & 0.33 & 0.64 & 0.41 & 0.92 & 0.62 & 0.36 & 0.87 \\
& RN18+RP-PCA    & 0.22 & 0.55 & 0.22 & 0.56 & 0.29 & 0.89 & 0.59 & 0.35 & 0.85 \\
& RN18+RP-MSM     & 0.34 & 0.62 & 0.34 & 0.63 & 0.41 & 0.89 & 0.68 & 0.38 & 0.89 \\
& RN18+MWT-PCA    & 0.50 & 0.76 & 0.50 & 0.77 & 0.55 & 0.94 & 0.84 & 0.42 & 0.95 \\
& RN18+MWT-MSM    & \textbf{0.53} & 0.75 & \textbf{0.53} & 0.76 & \textbf{0.58} & 0.92 & \textbf{0.86} & \textbf{0.45} & \underline{0.95} \\
& RN18+RWT-PCA    & 0.48 & 0.73 & 0.48 & 0.74 & 0.54 & 0.92 & 0.81 & 0.40 & 0.93 \\
& RN18+RWT-MSM    & 0.51 & 0.74 & 0.50 & 0.75 & 0.57 & 0.92 & 0.83 & 0.42 & 0.94 \\
\hline
\multirow{24}{*}{\rotatebox{90}{\quad \textbf{Baselines} \quad}}
& xLSTMAD           & 0.42 & \textbf{0.80} & 0.50 & \underline{0.85} & 0.51 & 0.70 & 0.59 & 0.44 & 0.86 \\

& CNN               & 0.36 & 0.75 & 0.35 & 0.77 & 0.41 & 0.80 & 0.68 & 0.41 & 0.88 \\
& OmniAnomaly       & 0.28 & 0.66 & 0.33 & 0.70 & 0.33 & 0.55 & 0.41 & 0.39 & 0.81 \\
& PCA               & 0.36 & 0.73 & 0.36 & 0.78 & 0.43 & 0.88 & 0.67 & 0.32 & 0.87 \\
& LSTMAD            & 0.36 & 0.73 & 0.36 & 0.77 & 0.41 & 0.83 & 0.68 & 0.43 & 0.89 \\
& USAD              & 0.27 & 0.65 & 0.32 & 0.69 & 0.32 & 0.53 & 0.40 & 0.39 & 0.80 \\
& AutoEncoder       & 0.36 & 0.70 & 0.35 & 0.72 & 0.40 & 0.64 & 0.51 & 0.33 & 0.82 \\
& KMeansAD          & 0.28 & 0.70 & 0.33 & 0.74 & 0.35 & 0.70 & 0.54 & 0.36 & 0.83 \\
& CBLOF             & 0.33 & 0.69 & 0.32 & 0.72 & 0.37 & 0.71 & 0.53 & 0.36 & 0.83 \\
& MCD               & 0.32 & 0.71 & 0.32 & 0.75 & 0.39 & 0.53 & 0.39 & 0.23 & 0.80 \\
& OCSVM             & 0.27 & 0.63 & 0.31 & 0.70 & 0.32 & 0.55 & 0.48 & 0.35 & 0.82 \\
& Donut             & 0.24 & \underline{0.78} & 0.32 & \textbf{0.86} & 0.34 & 0.63 & 0.44 & 0.26 & \textbf{0.98} \\
& RobustPCA         & 0.28 & 0.60 & 0.28 & 0.63 & 0.33 & 0.62 & 0.46 & 0.37 & 0.82 \\
& FITS              & 0.17 & 0.60 & 0.24 & 0.69 & 0.24 & 0.70 & 0.32 & 0.17 & 0.82 \\
& OFA               & 0.14 & 0.57 & 0.22 & 0.65 & 0.21 & 0.73 & 0.39 & 0.15 & 0.83 \\
& EIF               & 0.22 & 0.69 & 0.25 & 0.73 & 0.30 & 0.77 & 0.51 & 0.30 & 0.83 \\
& COPOD             & 0.23 & 0.68 & 0.23 & 0.72 & 0.31 & 0.76 & 0.47 & 0.27 & 0.82 \\
& IForest           & 0.22 & 0.68 & 0.23 & 0.71 & 0.29 & 0.71 & 0.44 & 0.26 & 0.81 \\
& HBOS              & 0.18 & 0.65 & 0.22 & 0.70 & 0.27 & 0.72 & 0.47 & 0.28 & 0.82 \\
& TimesNet          & 0.14 & 0.58 & 0.21 & 0.67 & 0.22 & 0.65 & 0.31 & 0.18 & 0.83 \\
& KNN               & 0.16 & 0.54 & 0.21 & 0.63 & 0.22 & 0.73 & 0.52 & 0.24 & 0.81 \\
& TranAD            & 0.15 & 0.60 & 0.20 & 0.66 & 0.23 & 0.68 & 0.40 & 0.22 & 0.79 \\
& LOF               & 0.11 & 0.54 & 0.16 & 0.62 & 0.17 & 0.60 & 0.36 & 0.16 & 0.77 \\
& AnomalyTransformer& 0.08 & 0.52 & 0.14 & 0.58 & 0.13 & 0.51 & 0.32 & 0.14 & 0.73 \\
\hline
\end{tabular}
\end{table*}

\subsection{TS2I projection schemes comparison}
Table~\ref{tab:ts2i-projection-schemes} compares the 15 TS2I projection schemes across detection performance and computational efficiency. RN-REP serves as a negative control, confirming that meaningful structure in the image representation is necessary for effective anomaly detection -- random pixel assignments yield near-chance performance.

Wavelet-based projection schemes achieved the strongest performance, with MWT-MSM reaching VUS-PR of 0.52. This superiority suggests that multi-scale temporal decomposition effectively preserves anomaly-discriminative patterns across different frequency bands. Interestingly, replication-based projection schemes (SG-REP, LG-REP) achieved remarkably competitive performance without requiring complex channelization. SG-REP is particularly compelling: achieving VUS-PR of 0.51 while being 7.8 times faster than LG-REP (1.40s) and 2.1 times faster than wavelet-based competitors (0.37s), it demonstrates that preserving the full multivariate structure through direct spatial mapping can be both effective and efficient. LG-REP achieved VUS-PR of 0.48 but at considerable computational expense (3.8 times slower than MWT-MSM), as its grid-based plotting incurs high computational overhead for dense line rendering.

These results reveal a nuanced performance-efficiency landscape with clear implications. MWT-MSM emerges as the optimal choice for accuracy-critical applications, accepting 2.1 times longer computation than SG-REP to gain 2.0\% VUS-PR improvement. Conversely, SG-REP may offer the best solution for real-time deployments, sacrificing detection quality for substantial speed gains \textbf{(RQ2)}. These results demonstrate that PRISM's modular design allows practitioners to select the optimal projection scheme for their deployment constraints -- from real-time applications favoring SG-REP to accuracy-critical settings benefiting from MWT-MSM -- without changing any other component of the pipeline.

\begin{table*}[h!]
\centering
\scriptsize
\caption{Mean performance across 9 evaluation metrics and computational time (seconds per 100 images) of 15 PRISM's TS2I projection schemes. Best results in \textbf{bold}, second-best \underline{underlined}.}
\label{tab:ts2i-projection-schemes}
\begin{tabular}{|l|c|c|c|c|c|c|c|c|c|c|}
\hline
\textbf{Projection} & \textbf{AUC-} & \textbf{AUC-} & \textbf{VUS-} & \textbf{VUS-} & \textbf{S-F1} & \textbf{PA-F1} & \textbf{E-F1} & \textbf{R-F1} & \textbf{A-F1} & \textbf{Time} \\
\textbf{scheme} & \textbf{PR} & \textbf{ROC} & \textbf{PR} & \textbf{ROC} & & & & & & \\
\hline
RN-REP      & 0.14 & 0.50 & 0.15 & 0.52 & 0.21 & 0.94 & 0.34 & 0.32 & 0.77 & \underline{0.20} \\
SG-REP      & \underline{0.51} & 0.76 & \underline{0.51} & 0.77 & 0.56 & 0.92 & 0.81 & 0.41 & 0.93 & \textbf{0.18} \\
LG-REP      & 0.49 & \textbf{0.78} & 0.48 & \textbf{0.79} & 0.53 & \textbf{0.95} & 0.79 & 0.39 & 0.93 & 1.40 \\

GASF-PCA    & 0.26 & 0.64 & 0.26 & 0.65 & 0.34 & \underline{0.95} & 0.68 & 0.37 & 0.89 & 0.32 \\
GASF-MSM    & 0.38 & 0.72 & 0.37 & 0.73 & 0.44 & 0.94 & 0.76 & 0.36 & 0.92 & 0.32 \\
GADF-PCA    & 0.29 & 0.59 & 0.28 & 0.61 & 0.36 & 0.93 & 0.67 & 0.38 & 0.89 & 0.34 \\
GADF-MSM    & 0.29 & 0.57 & 0.28 & 0.59 & 0.38 & 0.93 & 0.66 & 0.39 & 0.88 & 0.33 \\
MTF-PCA     & 0.23 & 0.58 & 0.23 & 0.59 & 0.30 & 0.93 & 0.60 & 0.37 & 0.86 & 0.32 \\
MTF-MSM     & 0.33 & 0.63 & 0.33 & 0.64 & 0.41 & 0.92 & 0.63 & 0.37 & 0.87 & 0.32 \\
RP-PCA      & 0.22 & 0.55 & 0.22 & 0.57 & 0.29 & 0.88 & 0.58 & 0.33 & 0.85 & 0.25 \\
RP-MSM      & 0.34 & 0.62 & 0.34 & 0.63 & 0.40 & 0.87 & 0.66 & 0.36 & 0.89 & 0.24 \\
MWT-PCA     & 0.50 & 0.76 & 0.50 & \underline{0.77} & 0.54 & 0.93 & 0.82 & 0.42 & 0.94 & 0.40 \\
MWT-MSM     & \textbf{0.53} & \underline{0.76} & \textbf{0.52} & 0.76 & \textbf{0.58} & 0.92 & \textbf{0.86} & \textbf{0.45} & \textbf{0.95} & 0.37 \\
RWT-PCA     & 0.47 & 0.73 & 0.47 & 0.74 & 0.53 & 0.92 & 0.79 & 0.40 & 0.93 & 0.39 \\
RWT-MSM     & 0.50 & 0.74 & 0.50 & 0.75 & \underline{0.56} & 0.92 & 0.83 & \underline{0.42} & \underline{0.94} & 0.37 \\
\hline
\end{tabular}
\end{table*}




\subsection{Channelization strategy: MSM vs PCA}
\label{sec:channelization}

To evaluate how channelization strategy affects detection quality, we compared MSM against PCA across all projection schemes for univariate-based TS2I transformations, models, and datasets. MSM outperformed PCA across the vast majority of configurations, achieving 18.9\% higher VUS-PR overall (0.390 vs 0.328). This advantage remained substantial for both CNN-AE (19.6\% improvement) and ResNet-AE (18.7\% improvement), indicating MSM's superiority is architecture-independent. Notably, MSM was also computationally 3.2\% faster than PCA channelization.

The impact of channelization varied substantially across TL strategies. FE showed the highest sensitivity to channelization choice, with MSM achieving 26.6\% higher VUS-PR than PCA (0.388 vs 0.306). This substantial gap suggests that when encoder weights remain fixed, the quality of input representation becomes critical -- MSM's deviation-sensitive statistics -- particularly standard deviation and maximum -- are theoretically better aligned with anomaly discriminability than PCA's variance-maximizing projection, consistent with the motivation in Section \ref{sec:univar-ts2i-projection-scheme}. PEU reduced this gap to 19.9\% (0.396 vs 0.330), as gradual adaptation partially compensates for weaker initial representations. DLR showed the smallest gap at 11.0\% (0.399 vs 0.359), indicating that fine-grained tuning can overcome limitations of suboptimal channelization.

We observe that MSM's advantage is most pronounced when model capacity to learn from data is constrained -- either by simpler architecture (CNN-AE) or frozen weights (FE strategy). Conversely, more flexible training regimes (DLR) reduce dependence on channelization quality. The consistent 11--27\% VUS-PR improvement across all configurations establishes MSM as the preferred channelization strategy for TSAD, with MSM-based projection schemes recommended for production deployments \textbf{(RQ3)}.

\subsection{Transfer learning from ImageNet: strategies and trade-offs}

Despite the substantial domain gap between natural images (ImageNet) and TS2I representations, TL from ImageNet-pretrained ResNet-AE consistently benefits TSAD across all strategies (Table~\ref{tab:transfer-learning}). 

Training only the decoder while keeping pretrained weights fixed (FE) achieved strong performance (VUS-PR: 0.35) with fastest training (94.21s), demonstrating that ImageNet features capture anomaly-relevant patterns without encoder adaptation. DLR reached the highest performance (VUS-PR: 0.38), only 8.6\% better than FE but requiring 77.5\% longer training (167.29s). PEU occupied the middle ground (VUS-PR: 0.37, 49.3\% overhead vs FE).

The seemingly narrow 8.6\% performance gap translates to meaningful differences in critical applications. In applications where missed anomalies carry high operational cost, even small gains in VUS-PR may justify additional training time. The trade-off is primarily computational: FE offers rapid deployment with 92\% of DLR's quality, while DLR maximizes accuracy at nearly doubled training cost. For fast deployment or resource-constrained environments, FE is optimal; for safety-critical or accuracy-demanding applications where the 8.6\% improvement justifies computational expense, DLR is preferable \textbf{(RQ4)}.

\begin{table}[h!]
\centering
\scriptsize
\caption{Mean performance across 9 evaluation metrics and training time (seconds) of 3 TL strategies. Best results in \textbf{bold}, second-best \underline{underlined}.}
\label{tab:transfer-learning}
\begin{tabular}{|l|c|c|c|c|c|c|c|c|c|c|}
\hline
\textbf{Strategy} & \textbf{AUC-PR} & \textbf{AUC-ROC} & \textbf{VUS-PR} & \textbf{VUS-ROC} & \textbf{S-F1} & \textbf{PA-F1} & \textbf{E-F1} & \textbf{R-F1} & \textbf{A-F1} & \textbf{Time} \\
\hline
FE & 0.36 & 0.66 & 0.35 & 0.67 & 0.42 & 0.93 & 0.70 & 0.38 & 0.89 & \textbf{94} \\
PEU & \underline{0.37} & \underline{0.67} & \underline{0.37} & \underline{0.68} & \underline{0.43} & \underline{0.93} & \underline{0.71} & \underline{0.38} & \underline{0.90} & \underline{141} \\
DLR & \textbf{0.38} & \textbf{0.68} & \textbf{0.38} & \textbf{0.69} & \textbf{0.44} & \textbf{0.93} & \textbf{0.71} & \textbf{0.39} & \textbf{0.90} & 167 \\
\hline
\end{tabular}
\end{table}


\subsection{Empirical time complexity}

PRISM's computational cost comprises TS2I transformation and model training. Using stride $s=1$, time-domain windowing generates 1k–400k images per dataset. Transformation scales linearly at 0.18–1.40s per 100 images but represents only 2–8\% of total time. Training time, which dominates the budget, depends on projection scheme and architecture choice. On average, the CNN-AE converges more slowly (2.10 min) than the deeper frozen ResNet-AE (1.57 min), while different TL strategies add varying overhead: PEU requires 2.35 min (+49\%) and DLR 2.79 min (+78\%). Projection schemes also impact efficiency -- models trained on TS2I from wavelets converge faster (MWT: 1.41 min, RWT: 1.55 min) than polar encodings (GASF: 3.05 min, GADF: 3.17 min). For median-sized datasets, end-to-end workflow time ranges 2.0–4.7 min. The transformation overhead is justified -- configurations that achieve 41\% average VUS-PR improvement over baselines require 2–5 min end-to-end \textbf{(RQ5)}.





\section{Conclusion}

We introduced PRISM, a systematic meta-workflow designed for multivariate TSAD through TS2I representations. PRISM's plug-and-play design enables any combination of TS2I transformation, channelization strategy, and detector to be systematically assessed. Evaluating 15 projection schemes across 14 datasets from the TSB-AD benchmark, configurations enabled by PRISM achieved the best performance on 10 of 14 datasets and remained competitive with 24 specialized time-domain baselines overall. Among key findings, we observe that: \romannumeral 1)~wavelet-based projection schemes with MSM channelization achieve best performance (MWT-MSM: 0.52 VUS-PR), with MSM providing 11--27\% gains over PCA; \romannumeral 2)~MSM's advantage peaks when model capacity is constrained (frozen encoder: +26.6\% vs fine-tuned: +11.0\%), demonstrating that superior preprocessing compensates for architectural limitations; \romannumeral 3)~despite the substantial domain gap, ImageNet features transfer effectively to TSAD, with frozen encoders retaining 92\% of fine-tuned performance while training 1.8 times faster.
In the future, we aim to study the impact of image resolution on detection performance, explore continual learning scenarios, and leverage data-driven window selection.

\begin{credits}
\subsubsection{\ackname} The research presented in this paper was supported by funds from: i) the Polish Ministry of Science and Higher Education allocated to the AGH University of Krakow; ii) SOCCER project (grant agreement no. 101128073), with the support of the European Cybersecurity Competence Centre (ECCC), and co-funded by the Polish Ministry of Science and Higher Education under the programme "Co-funded International Projects."
\end{credits}

%
%
%
%

\bibliographystyle{splncs04}
\bibliography{bibliography}

\end{document}